%% file: document.tex
\documentclass[letter, 10pt, conference]{ieeeconf}     
                                                        
\IEEEoverridecommandlockouts                              
                                                          
\usepackage{cite}      
\usepackage{graphicx}  
\usepackage{subfigure} 
\usepackage{url}       
\usepackage{amsmath}  
\usepackage{amssymb}
\usepackage{epsfig}     
\usepackage{color}
\usepackage[mathcal]{euscript}
\usepackage{mathrsfs}
\usepackage[ruled,vlined,linesnumbered]{algorithm2e}
\usepackage[scale={0.8,0.8}]{geometry}
\usepackage{rotating}
\usepackage{url}
\usepackage{multirow}
\usepackage{multicol}
\usepackage{eurosym}
\usepackage{framed}
\usepackage{footnote}
\usepackage{float}
\usepackage{caption}
\usepackage{fancyhdr}
\usepackage{paralist}
\usepackage{etoolbox}
\usepackage{listings}
\usepackage{textcomp}
\definecolor{listinggray}{gray}{0.95}
\definecolor{lbcolor}{rgb}{0.95,0.95,0.95}

\input{macroGeometricPrimitives.tex}

\input{macroSemantics.tex}

\input{macroSemanticOperations.tex}

\input{listingsSettings}

\graphicspath{{images/}{images_geometricSemantics/}{listings/}}

\title{\LARGE \bf Domain Specific Language for Geometric Relations between Rigid Bodies targeted to robotic applications} 

\author{Tinne De Laet, Wouter Schaekers, Jonas de Greef, and Herman Bruyninckx
\thanks{Tinne De Laet and Herman Bruyninckx are with the Department of Mechanical Engineering,
Katholieke Universiteit Leuven, Belgium. 
Wouter Schaekers and Jonas de Greef are students at the Computer Science Engineering Department,
Katholieke Universiteit Leuven, Belgium. 
Corresponding author: Tinne De Laet
(\protect\url{Tinne.DeLaet@mech.kuleuven.be})}%
}

\begin{document}
       
\maketitle
\thispagestyle{empty}
\pagestyle{empty}

\maketitle

\begin{abstract}
This paper presents a DSL for geometric relations between rigid bodies such as relative position, orientation, pose, linear velocity, angular velocity, and twist.
The DSL is the formal model of the recently proposed semantics for the standardization of geometric relations between rigid bodies~\cite{DeLaet-ram2012a,DeLaet-rrfc2012}, referred to as `geometric semantics'. This semantics explicitly states the
coordinate-invariant properties and operations, and, more importantly, all
the choices that are made in coordinate representations of these geometric
relations. This results in a set of concrete suggestions for standardizing
terminology and notation, allowing programmers to write fully unambiguous
software interfaces, including automatic checks for semantic correctness of
all geometric operations on rigid-body coordinate representations.

The DSL is implemented in two different ways:
an external DSL in Xcore and an internal DSL in Prolog.
Besides defining a grammar and operations, the DSL also implements constraints.
In the Xcore model, the Object Constraint Language language is used, while in the Prolog model, the constraint are natively modelled in Prolog.

This paper discusses the implemented DSL and the tools developed on top of this DSL. In particular an editor, checking the semantic constraints and providing semantic meaningful errors during editing is proposed.

\end{abstract}

\section{Introduction}
\label{sec:introduction}

When developing robotic applications, robot programmers and application developers have to deal with three-dimensional motion and relations between rigid bodies (manipulated objects, robot links, or mobile bases).
Rigid bodies are essential primitives in the modelling of robotic devices,
tasks and perception. Basic geometric relations between rigid bodies include
relative position, orientation, pose (combining position and orientation), linear velocity, angular
velocity, and twist (combining linear and angular velocity). 
To express geometric relations and perform mathematical operations on them (e.g. composition of relative motion, time differentiation, or integration), robot programmers have to choose coordinate representations with which to perform the corresponding numerical operations.

Until recently, and despite a history of over 50 years, the geometric properties of rigid-body operations, and their coordinate representations, were not standardized, which has led to a proliferation of mutually incompatible software libraries, in the robot control products of commercial manufacturers as well as in \emph{open source} libraries such as KDL (Kinematics and Dynamics Library)~\cite{kdl-url}, ROS (Robot Operating System)~\cite{ros-url}, RL (Robotics Library)~\cite{robotics-library-url}, \dots.
All geometric relations and their coordinate representations entail a surprisingly large number of choices or assumptions, which are often made implicitly, but which are necessary to consider in view of (i) understanding the physical meaning of the numerical values that constitute the coordinate representation of a geometric relation and (ii) performing physically meaningful mathematical operations on these numerical values.
Not explicitly stating the above assumptions may lead to errors in the calculations (composition of geometric relations expressed in different coordinate frames, composition of poses and orientation coordinate representations in wrong order,\ldots \cite{DeLaet-ram2012a}).
To alleviate this problem, we recently proposed semantics for the standardization of geometric relations between rigid bodies~\cite{DeLaet-ram2012a}, referred to as `geometric semantics'. This semantics explicitly states the
coordinate-invariant properties and operations, and, more importantly, all
the choices that are made in coordinate representations of these geometric
relations. This results in a set of concrete suggestions for standardizing
terminology and notation, allowing programmers to write fully unambiguous
software interfaces, including automatic checks for semantic correctness of
all geometric operations on rigid-body coordinate representations.
This resulted in a Robot Request for Comments~\cite{DeLaet-rrfc2012} for the Robot Engineering Task Force~\cite{retf-url}.
Furthermore, software providing a C++ implementation of the software is developed and available as open-source~\cite{DeLaet-ram2012b,GeometricSemanticsSoftware-url}.

Domain Specific Languages are lightweight programming languages designed to concisely express the concepts of a particular domain. Commonly two types are distinguished: internal and external DSLs. The former are built on top of an existing language, while the latter are developed from scratch resulting in a custom syntax and making them independent from existing languages.
By reusing existing infrastructure, internal DSLs are easier to create, maintain, and combine with other DSLs than external ones~\cite{Klotzbuecher2011-IROS}.
External DSLs, while suffering from an increased cost for creating and maintenance, are not constrained by any other language.
Therefore, the choice between an internal or external design often depends on the particular application, use case, available tools, and preferences of the designer.
In this paper we develop both types for the geometric semantics: \begin{inparaenum}
                                                                  \item an \emph{external} DSL in Xcore and
                                                                  \item an \emph{internal} DSL in Prolog.
                                                                 \end{inparaenum}

The goal of this paper is fourfold. Firstly, we want to build a DSL for geometric relations between rigid bodies such as relative position, orientation, pose, linear velocity, angular velocity, and twist founded on the geometric semantics~\cite{DeLaet-ram2012a,DeLaet-rrfc2012}.
This DSL advances with respect to the available available implementation in the general-purpose programming language C++, by formalizing the underlying model of the geometric semantics.
Furthermore, the DSL is the basis for the developments of tools that assist the robot programmers and application developers to write fully unambiguous
software interfaces and prevent common errors in geometric calculations. In particular this paper presents and editor built on top of the proposed DSL that automatically checks the semantic correctness of
all geometric operations on rigid-body coordinate representations, while writing and editing the code.
Secondly, we want to explore the impact of different design choices (internal, external), work flows, and tools.
Thirdly, we believe that due to the concise and mature nature of the underlying geometric semantics theory and its relevance for the robotics domain it will prove to be an excellent example for future DSL development in robotics. 
Fourthly, we will highlight the unfulfilled robotic needs still present in Model Driven Engineering.


%



Section~\ref{sec:related} gives an overview of related work.
Section~\ref{sec:geometricSemantics} provides a short summary of the geometric semantics theory relevant for this paper.
Section~\ref{sec:levelsAbstraction} situates this paper's contributions using the four levels of abstraction in Model Driven Engineering.

\section{Related work}
\label{sec:related}
Since we are not aware of any DSL on the semantics for geometric relationships between rigid bodies, our related work will rather point to some other DSLs developed in the robotics domain.

Frigerio et al.'s DSL is the DSL most related to the DSL proposed in this paper. They propose a DSL for kinematic models and fast implementation of robot dynamic algorithms.
The DSL allows to model algorithms that are parametrised on the kinematics/dynamics model of a robot, hereby facilitating the generation of executable code tailored for a specific robot. This approach only requires the users to provide a high level description of their robot and relieves them from hand-crafted development.
\newline
Furthermore, we want to mention the Mechatronics Description Language (MDL), which is a domain-specific language that can model the kinematic structure of individual robot modules and declaratively describe their possible interconnections. From this description, the MDL compiler generates the code that is needed to simulate the resulting robots within Webots, a widely used commercial robot simulator, and the software component needed for spatial structure computations by a virtual machine-based runtime system, which we have developed and use for programming physical modular robots~\cite{bordignon2010}. 

Klotzb\"ucher et al.~\cite{Klotzbuecher2010-DYROS} propose a DSL for specifying robotic tasks using the task frame formalism as an example of lua as a lightweight and composable metamodelling language for specification and validation of internal DSLs.
In later work Klotzb\"ucher et al.~\cite{Klotzbuecher2011-IROS}) propose a DSL allowing to separate task specification and coordination of these tasks using state charts.


\section{Levels of abstraction in Model Driven Engineering}
\label{sec:levelsAbstraction}
\begin{figure*}
\centering
	\resizebox{0.65\textwidth}{!}{\includegraphics{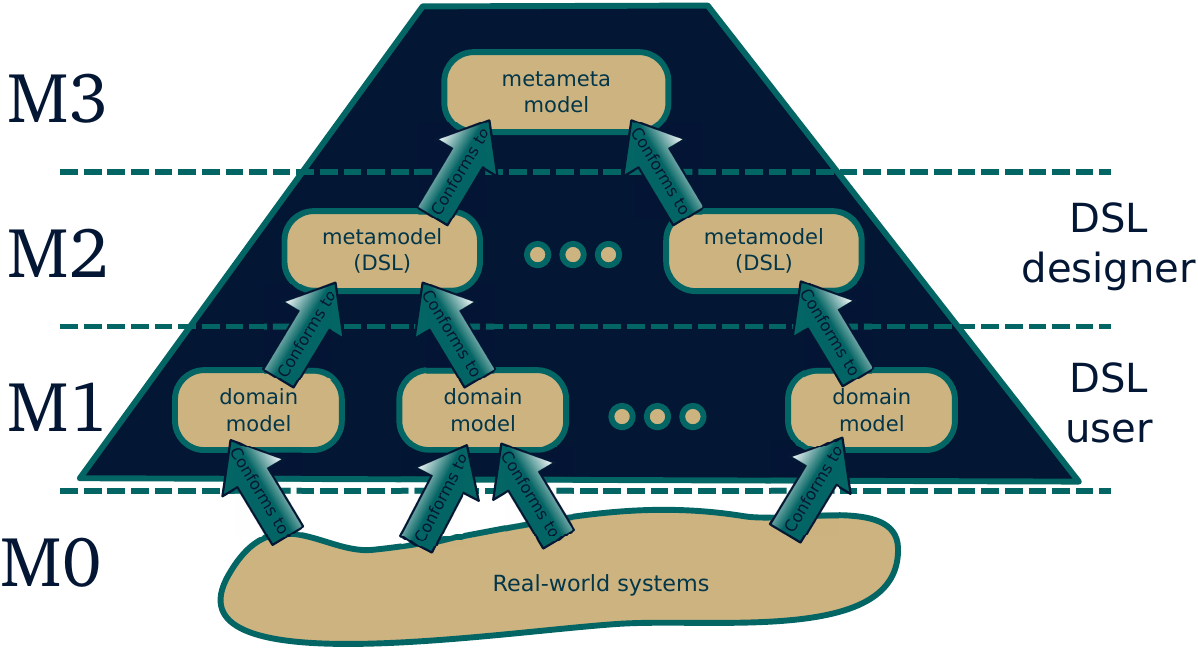}}
	\caption{The four levels of abstraction.)}
	\label{fig:levelsAbstraction}
\end{figure*}

Figure~\ref{fig:levelsAbstraction} illustrates a systematic approach to model a certain domain in \emph{four levels of abstraction} \cite{Bezivin2005,Klotzbuecher2011-IROS}. 
These four levels have the following meaning for the context of the geometric semantics:
\begin{itemize}
 \item[\textbf{M0:}] the level of the \emph{concrete implementations}, for instance a particular set of geometric semantics calculations using the C++ library of the geometric semantics~\cite{GeometricSemanticsSoftware-url},
 \item[\textbf{M1:}] the level of a \emph{particular set} of geometric semantics calculations using the geometric semantics DSL,
 \item[\textbf{M2:}] the level of the \emph{application independent geometric semantics DSL}, which provides a language for both coordinate representation independent and dependent (taking into account the constraints of a particular coordinate representation) geometric calculations.
 \item[\textbf{M3:}] the highest level of abstraction, that is, the model in terms of which we describe our meta-models (M2). For example, ecore that we can use to describe our geometric semantics DSL.
\end{itemize}

The geometric semantics theory~\cite{DeLaet-ram2012a}, summarized in Section~\ref{sec:geometricSemantics}, can be considered as the basis for the \textbf{M2} level DSL, as it describes (in language) the constraints on the geometric relations semantics, the possible operations on the geometric relations, the constraints on the relations and operations, and the constraints imposed by particular coordinate representations.
The available C++ implementation~\cite{GeometricSemanticsSoftware-url} and the applications implemented in it, are examples of the \textbf{M0} level.

This paper provides DSL implementations, both an external DSL and an internal DSL, on the \textbf{M2} level.
Furthermore, this paper presents tools based on the developed DSLs, that allow the DSL users to implement their particular set of geometric semantic calculations, i.e. to work on the \textbf{M1} level.
To illustrate the proposed approach, we provide an example on a M1 implementation for a particular geometric calculation and show how the developed DSL and the accompanying tools will help to prevent commonly made errors.

\section{Geometric semantics, background \cite{DeLaet-ram2012a}}
\label{sec:geometricSemantics}

\subsection{Geometric relations}
\label{subsec:geometricSemantics:geometricRelations}

\begin{figure}
\centering
	\resizebox{0.7\columnwidth}{!}{\includegraphics{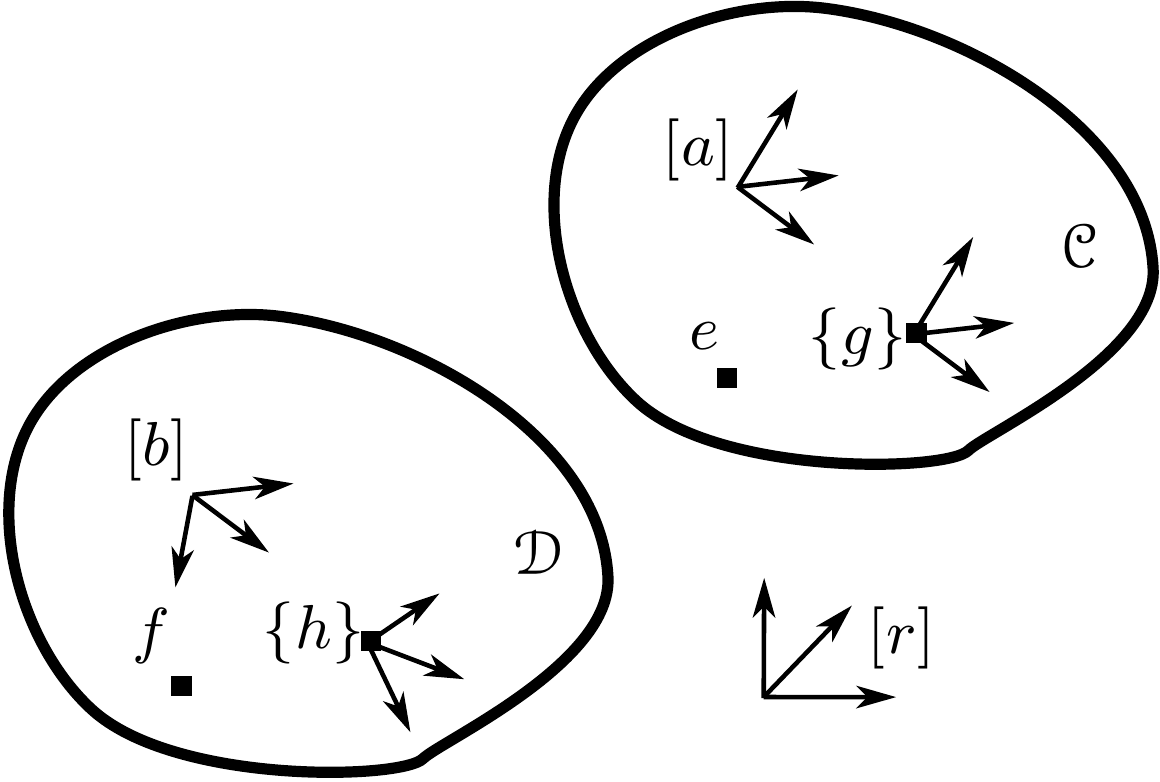}}
	\caption{The geometric relation between rigid bodies are described using a set of \emph{geometric primitives}: points, vectors, orientation frames, and frames. The above figure shows the geometric primitives that are useful to define the position, orientation, pose, linear velocity, angular velocity, and twist of body $\body{C}$ with respect to body $\body{D}$:
	an orientation frame $\protect\orientation{a}$, a point $\protect\point{e}$, and frame $\protect\refframe{g}$ fixed to body $\body{C}$, an orientation frame $\protect\orientation{b}$, a point $\protect\point{f}$, and frame $\protect\refframe{h}$ fixed to body $\body{D}$, and a coordinate frame $\protect\orientation{r}$, considered instantaneously fixed to body $\body{D}$, in which the coordinates are expressed. (Extract from~\cite{DeLaet-ram2012a}.)}
	\label{fig:primitves}
\end{figure}

Geometric relations between bodies are described using a set of \emph{geometric primitives}\footnote{This background contains a short summary of the semantics for the standardization of geometric relations between rigid bodies, for more details we refer to \cite{DeLaet-ram2012a}.}:
points ($\point{e}$), vectors, orientation frames ($\orientation{a}$ , they represent an orientation, by means of
three orthonormal vectors indicating the frame's X-axis $\axis{X}$, Y-axis
$\axis{Y}$, and Z-axis $\axis{Z}$), and frames ($\refframe{g}$). 
Figure~\ref{fig:primitves} presents the geometric primitives body, point, vector, orientation frame, and frame graphically. To help the reader we will consistently use the following naming for the geometric primitives to represent the geometric relation of a body $\body{C}$ with respect to body $\body{D}$ in this document:  $\fixedTo{\point{e}}{C}$, $\fixedTo{\orientation{a}}{C}$, $\fixedTo{\refframe{g}}{C}$, $\fixedTo{\point{f}}{D}$, $\fixedTo{\orientation{b}}{D}$, and $\fixedTo{\refframe{h}}{D}$.

Table~\ref{tab:semantics} summarizes the minimal but complete set of geometric primitives and the (coordinate) semantics for the geometric relations position, orientation, pose,  twist between rigid bodies, which are the most relevant relations for this paper. 

 \begin{table*}
  \begin{center}
   \begin{tabular*}{1\textwidth}{@{\extracolsep{\fill}}llllc} \hline
\textbf{Geometric Relation} & \textbf{(Coordinate) semantics} & \textbf{Geometric primitives} & \textbf{Graphical representation} \\ \hline \\
\textbf{Position}  & $\Position{e}{C}{f}{D}{}$  & point $\point{e}$ & \multirow{6}{*}{
\resizebox{0.12\textwidth}{!}{\includegraphics{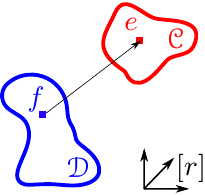}}
}\\ 
& $\Position{e}{C}{f}{D}{r}$		& body $\body{C}$ \\
&& reference point $\point{f}$ \\
&& reference body $\body{D}$ \\
&& coordinate frame $\orientation{r}$ \\ 
 \multicolumn{3}{p{15cm}}{\emph{Position of point $\point{e}$ fixed to body $\body{C}$ ($\fixedTo{\point{e}}{C}$) with respect to 
point $\point{f}$ fixed to body $\body{D}$ ($\fixedTo{\point{f}}{D}$), expressed in coordinate frame $\orientation{r}$}}
\\
 \\
\textbf{Orientation}  & $\Orientation{a}{C}{b}{D}{}$  & orientation frame $\orientation{a}$ 
& \multirow{3}{*}{
\resizebox{0.12\textwidth}{!}{\includegraphics{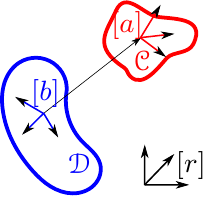}}
}
\\
& $\Orientation{a}{C}{b}{D}{r}$ & body $\body{C}$ \\
&& reference orientation frame $\orientation{b}$ \\
&& reference body $\body{D}$ \\
&& coordinate frame $\orientation{r}$ \\
 \multicolumn{3}{p{15cm}}{\emph{Orientation of orientation frame $\orientation{a}$ fixed to body $\body{C}$ ($\fixedTo{\orientation{a}}{C}$) with respect to 
orientation frame $\orientation{b}$ fixed to body $\body{D}$ ($\fixedTo{\orientation{b}}{D}$), expressed in coordinate frame $\orientation{r}$}}
\\
\\
\textbf{Pose}  & $\Pose{e}{a}{C}{f}{b}{D}{}$  & point $\point{e}$ 
& \multirow{3}{*}{
\resizebox{0.12\textwidth}{!}{\includegraphics{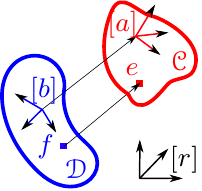}}
}
\\
& $\Pose{e}{a}{C}{f}{b}{D}{r}$ & orientation frame $\orientation{a}$ \\
&& body $\body{C}$ \\
&& reference point $\point{f}$ \\
&& reference orientation frame $\orientation{b}$ \\
&& reference body $\body{D}$ \\
&& coordinate frame $\orientation{r}$ \\ 
\\
\multicolumn{3}{p{15cm}}{\emph{Pose of point $\point{e}$ and orientation frame $\orientation{a}$ fixed to body $\body{C}$ ($\fixedTo{(\point{e},\orientation{a})}{C}$) with respect to 
point $\point{f}$ and orientation frame $\orientation{b}$ fixed to body $\body{D}$ ($\fixedTo{(\point{f},\orientation{b})}{D}$), expressed in coordinate frame $\orientation{r}$}}
\\
\\

& $\Pose{g}{g}{C}{h}{h}{D}{}$    & frame $\refframe{g}$ 
& \multirow{6}{*}{
\resizebox{0.12\textwidth}{!}{\includegraphics{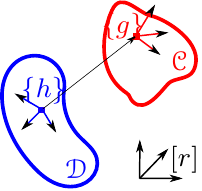}}
}
\\
& $\Pose{g}{g}{C}{h}{h}{D}{r}$ & body $\body{C}$ \\
&& frame $\refframe{h}$ \\
&& reference body $\body{D}$ \\ 
&& coordinate frame $\orientation{r}$ \\
\multicolumn{3}{p{15cm}}{\emph{Pose of frame $\refframe{g}$ fixed to body $\body{C}$ ($\fixedTo{\refframe{g}}{C}$) with respect to 
frame $\refframe{g}$ fixed to body $\body{D}$ ($\fixedTo{\refframe{g}}{D}$), expressed in coordinate frame $\orientation{r}$}}
\\ \\
\\
\textbf{Linear velocity }& $\LinearVelocity{e}{C}{}{D}{}$ & point $\point{e}$ 
& \multirow{6}{*}{
\resizebox{0.12\textwidth}{!}{\includegraphics{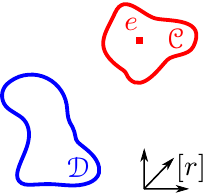}}
} 
\\
&  $\LinearVelocity{e}{C}{}{D}{r}$ & body $\body{C}$ \\
&& reference body $\body{D}$ \\
&& coordinate frame $\orientation{r}$ \\
\multicolumn{3}{p{15cm}}{\emph{Linear velocity of point $\point{e}$ fixed to body $\body{C}$ ($\fixedTo{\point{e}}{C}$) with respect to body $\body{D}$, expressed in coordinate frame $\orientation{r}$}}
\\ \\ 
\\
\textbf{Angular velocity} & $\AngularVelocity{}{C}{}{D}{}$ & body $\body{C}$
& \multirow{6}{*}{
\resizebox{0.12\textwidth}{!}{\includegraphics{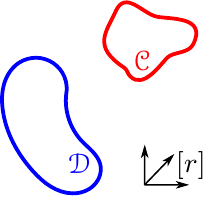}}
} 
\\
 &    $\AngularVelocity{}{C}{}{D}{r}$ & reference body $\body{D}$ \\
&& coordinate frame $\orientation{r}$ \\
\multicolumn{3}{p{15cm}}{\emph{Angular velocity of body $\body{C}$ with respect to body $\body{D}$, expressed in coordinate frame $\orientation{r}$}}
\\
\\
\textbf{Twist} & $\Velocity{e}{}{C}{}{}{D}{}$  & point $\point{e}$ 
& \multirow{6}{*}{
\resizebox{0.12\textwidth}{!}{\includegraphics{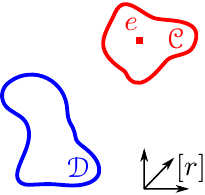}}
} 
\\
&    $\Velocity{e}{}{C}{}{}{D}{r}$ & body $\body{C}$ \\
&& reference body $\body{D}$ \\
&& coordinate frame $\orientation{r}$ \\
\multicolumn{3}{p{15cm}}{\emph{Twist of body $\body{C}$ with velocity reference point $\point{e}$ ($\fixedTo{\point{e}}{C}$) with respect to body $\body{D}$, expressed in coordinate frame $\orientation{r}$}}
\\
\\ \hline
   \end{tabular*}
     \end{center}
\caption{Minimal semantics and coordinate semantics (expressed in coordinate frame $\protect\orientation{r}$) 
including the minimal but complete set of geometric primitives for the position, orientation, pose, linear velocity, angular velocity, and twist of body $\body{C}$ with point $\protect\point{e}$, orientation frame $\protect\orientation{a}$, and frame $\protect\refframe{g}$ with respect to $\body{D}$ with point $\protect\point{f}$, orientation frame $\protect\orientation{b}$, and frame $\protect\refframe{h}$, including a graphical representation. (extracted from~\cite{DeLaet-ram2012a})}
\label{tab:semantics}
  \end{table*}

\subsection{Semantic operations}
\label{subsec:geometricSemantics:semanticOperations}
  
On the geometric relations defined in Section~\ref{subsec:geometricSemantics:geometricRelations}, semantic operations that compose the geometric relations or that change the point, orientation frame, reference point, reference orientation frame, or coordinate frame of the geometric relation can be applied.
These semantic operations themselves impose constraints on the geometric relation they are applied to and on the operation arguments (which are themselves geometric relations) of the operator.
While \cite{DeLaet-ram2012a} provides an overview of semantic operations that can be applied to geometric relations and lists the constraints imposed by the operations, we will only give an example illustrating the concept of the semantic operation and the constraints imposed by it.

As an example, consider the semantic operation to change the point used to describe the position of body $\body{C}$ with respect to body $\body{D}$. 
Imagine $\Position{e_1}{C}{f}{D}{r}$ is the semantic description of the position of body $\body{C}$ with respect to body $\body{D}$. To change the point to describe the position from the current point $\point{e_1}$ to a new point $\point{e_2}$, the position of the new point $\point{e_2}$ fixed to body $\body{C}$ with respect to $\point{e_1}$ fixed to body $\body{C}$ and expressed in the same coordinate frame $\orientation{r}$ is needed, i.e. $\Position{e_2}{C}{e_1}{C}{r}$.
If the semantic operator $\changePointPosition{}{}$ is applied to the geometric relation of which the point has to be changed (in our example $\Position{e_1}{C}{f}{D}{r}$) and has as an argument the geometric relation needed to achieve this change of reference point, the $\changePointPosition{}{}$ imposes the following constraints: (1) the argument of $\changePointPosition{}{}$ should be a PositionCoord geometric relation; (2) the reference point of the argument should be equal to the point of the position the operator is applied on; (3) the body of the argument should be equal to the body of the position the operator is applied on;
(4) the reference body of the argument should be equal to body of the position the operator is applied on; and (5) the coordinate frame of the argument should 
be equal to the point of the position the operator is applied on.
This can be visually illustrated as follows:
\begin{multline}
\Position{e_2}{C}{f}{D}{r} =\\ {\Position{\underline{e_1}}{\underline{\underline{C}}}{f}{D}{\underline{\underline{\underline{r}}}}} .\textrm{changePointPosition}( \\
{\Position{e_2}{\underline{\underline{C}}}{\underline{e_1}}{\underline{\underline{C}}}{\underline{\underline{\underline{r}}}}}),
\end{multline}
the semantic
constraints imposed on the geometric relation the operation is applied to, and on the operation arguments, are shown by using the same names for the geometric primitives when equality of the primitives is imposed, furthermore the lines indicate ones again the geometric primitives that should be equal to obtain a semantically correct operation.

\section{Geometric semantics DSL (M2) and the tooling (M1)}
\label{sec:geometricSemanticsDSL}

\subsection{External DSL}
\label{subsec:geometricSemanticsDSL:external}
\subsubsection{DSL design}
The external DSL is developed using Xcore. As this DSL is not using the java specific syntax parts of Xcore, it can be considered as a plain text file and therefore as an external DSL.
The geometric semantics DSL uses the Object Contraint language (OCL) DSL to define the constraints in the geometric semantics. Defining these constraints only requires a small set of constraints of OCL, making it feasible (although not necessarily desired) to eliminate the dependency on OCL with limited effort.

Next we discuss the design of the DSL in Xcore.
Our DSL consists of a `root' class called `DomainModel'.
This DomainModel class contains DomainRules. A DomainRule consists of `Primitive', `GeometricRelation', `GeometricCoordinateRelation', `SemanticOperation'
and `SemanticCoordinateOperation'.
Listing~1 shows the definition of the primitive Point and the geometric (coordinate) relation PositionSemantics and PositionCoordinateSemantics.

\begin{figure*}
 \resizebox{\textwidth}{!}{
 \centering
 \includegraphics{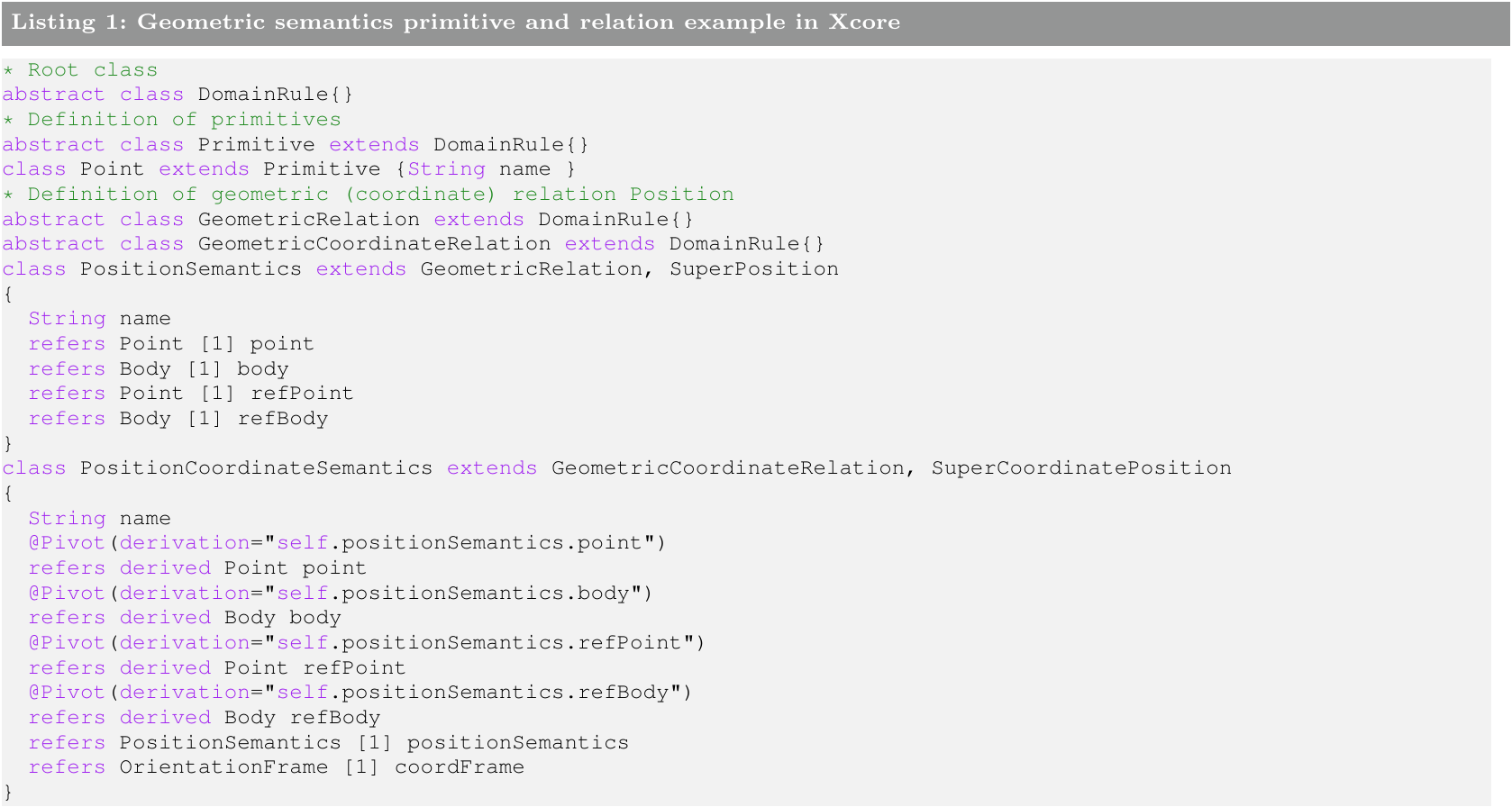}
}
\end{figure*}

%
%

Listing~2 shows the definition of the PositionChangePoint geometric operation and the constraints (defined using OCL) to which this operation has to comply.

\begin{figure*}
 \resizebox{\textwidth}{!}{
 \centering
 \includegraphics{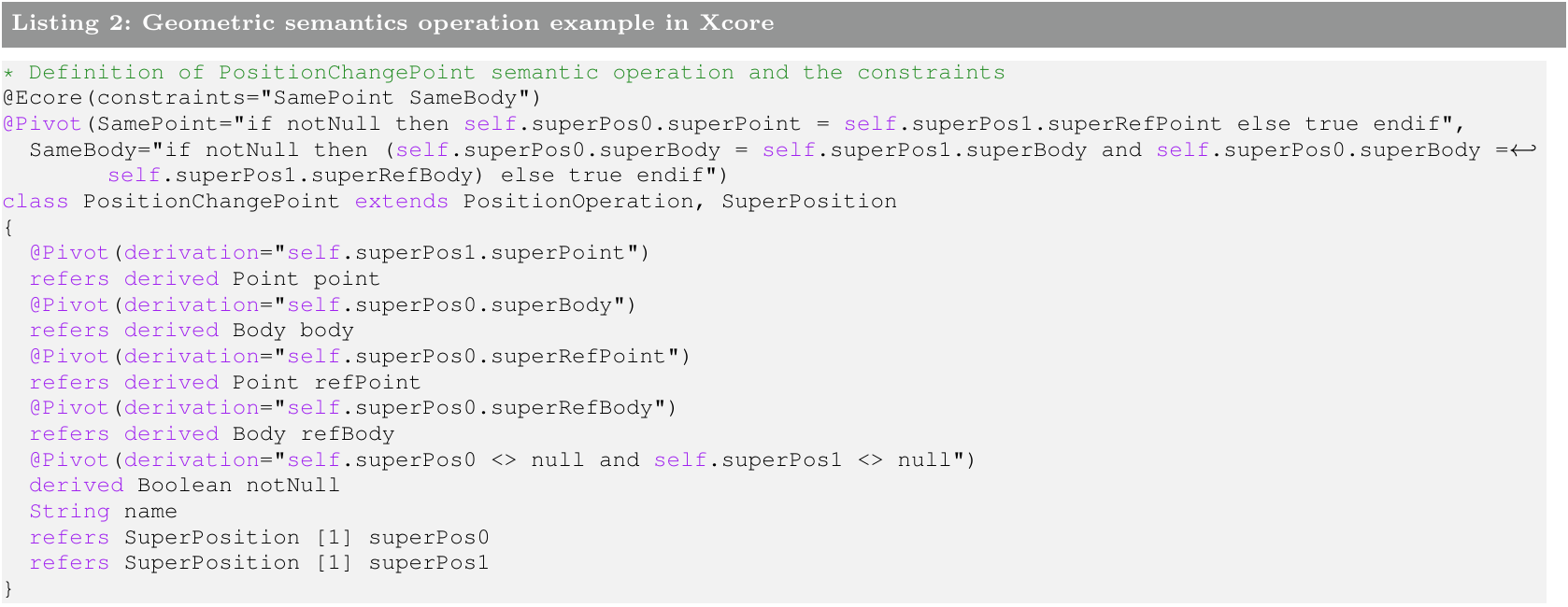}
}
\end{figure*}

\subsubsection{Work flow and tooling}
Thanks to the Xcore support in DSL we can make use of a Model Driven Engineering work flow in Eclipse.
The DSL can be loaded inside Eclispe and converted into an ecore model and subsequently in an Xtext model. The tooling (such as an editor) made available by Xtext can be created to support the M1 level for the geometric semantics.
The Xtext editor hereby allows for semantic checking of the geometric semantics during editing, hereby reducing application development time since errors are detected very early.

\subsection{Internal DSL}
\label{subsec:geometricSemanticsDSL:internal}
\subsubsection{DSL design}
The internal DSL is built on top of Prolog. 
This way Prolog can be used to define the grammar, and in particular the logic constraints of the geometric semantics.
Furthermore it provides a good mechanism to provide bookkeeping of the geometric primitives and relations in a particular application (which points, orientation frames, poses, \ldots are defined and check if they are uniquely defined).
Listing~3 shows the definition of the geometric (coordinate) relation Position (defining both PositionSemantics and PositionCoordinateSemantics).

\begin{figure*}
 \resizebox{\textwidth}{!}{
 \centering
 \includegraphics{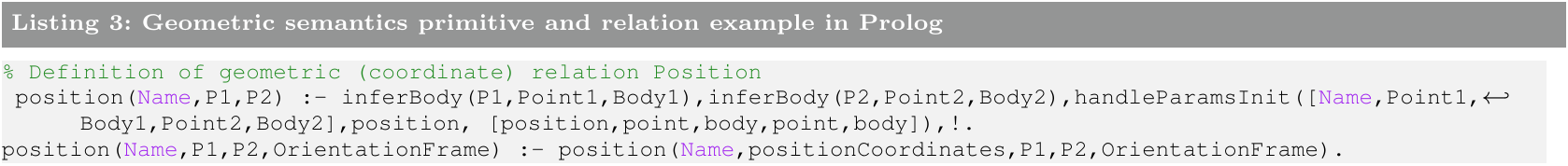}
}
\end{figure*}


Listing~4 shows the definition of the PositionChangePoint geometric operation and the constraints to which this operation has to comply.

\begin{figure*}
 \resizebox{\textwidth}{!}{
 \centering
 \includegraphics{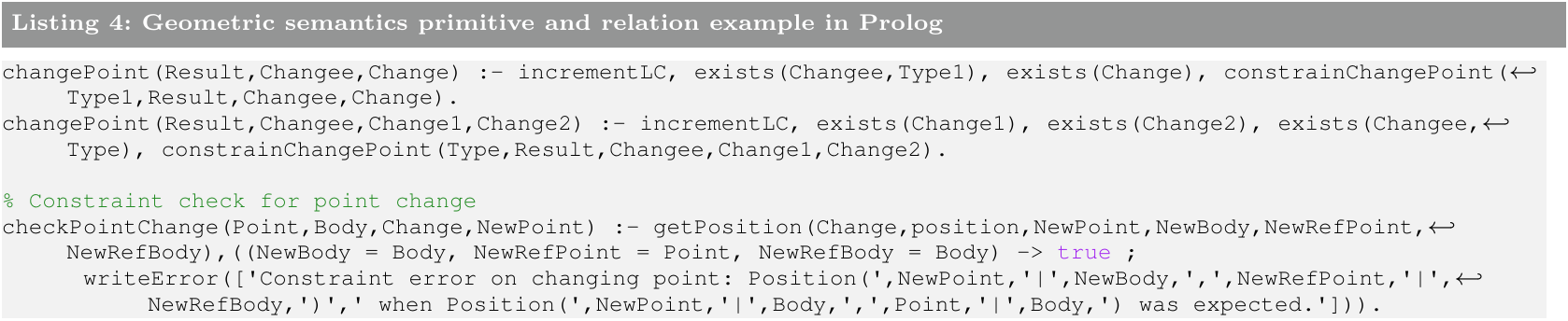}
}
\end{figure*}

%

\subsubsection{Work flow and tooling}
The work flow is a particular work flow based on the Prolog language. As a tool we developed our own editor that allows the DSL user to use the syntax as proposed in the geometric semantics theory~\cite{DeLaet-ram2012a}, but parses it to Prolog code which is subsequently executed in the background to do the semantic checking. This way, similar functionality as obtained with the Xtext editor is obtained, i.e. semantic checking is done and meaningful error statements are produced during editing.

%

\section{Example}
\label{sec:example}
\begin{figure*}
\centering
	\resizebox{1\textwidth}{!}{\includegraphics{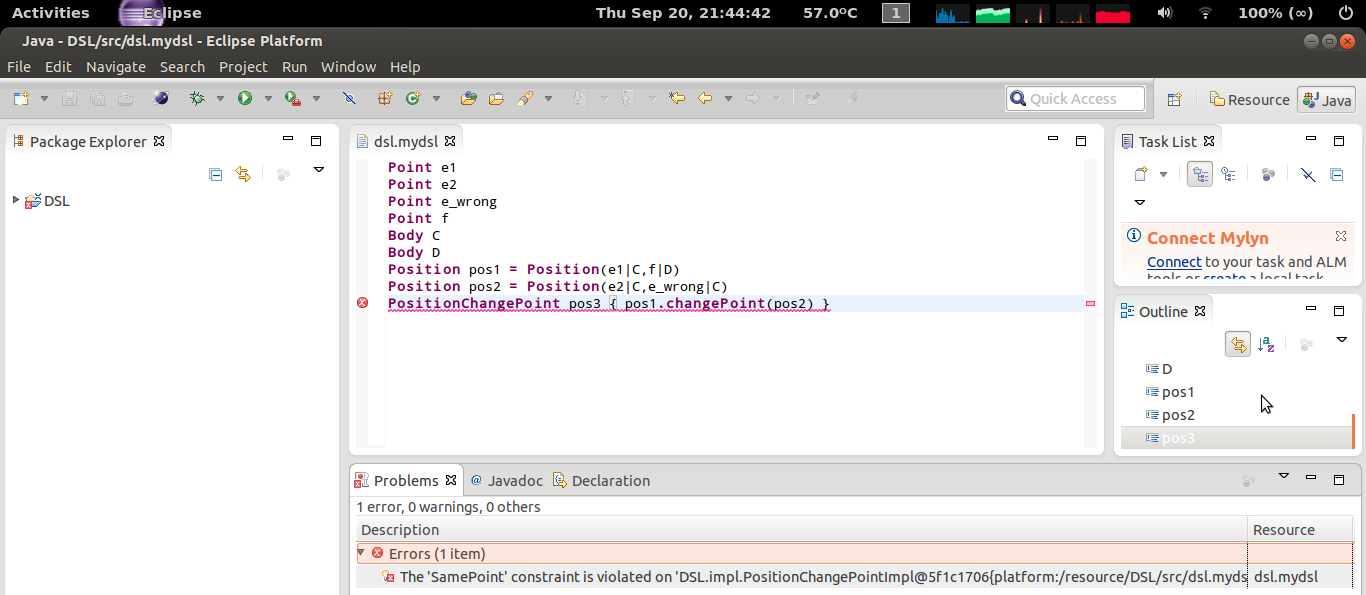}}
	\caption{Xtext editor example when violating the constraint that the point of the subject has to be equal to the reference point of the argument when applying the geometric operation changePoint.}
	\label{fig:errorXtextEditor}
\end{figure*}

\begin{figure*}
\centering
	\resizebox{1\textwidth}{!}{\includegraphics{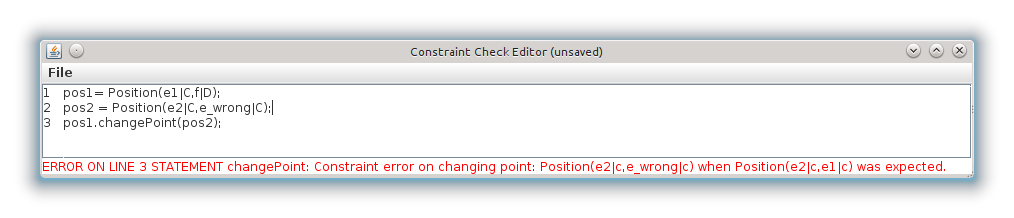}}
	\caption{Prolog-based editor example when violating the constraint that the point of the subject has to be equal to the reference point of the argument when applying the geometric operation changePoint.}
	\label{fig:errorPrologEditor}
\end{figure*}

This section illustrates how the DSL can be used to prevent common errors in geometric calculations.
To this end we use the following semantic operations:
\begin{multline}
{\Position{\underline{e_1}}{\underline{\underline{C}}}{f}{D}{}} .\textrm{changePointPosition}( \\
{\Position{e_2}{\underline{\underline{C}}}{\underline{e_1}}{\underline{\underline{C}}}{}}).
\end{multline}
In the above statement the lines illustrate the constraints on the semantic operation i.e. (we refer to the PositionCoord to which the operator is applied to as the subject and to the PositionCoord that is used as an argument in the operation as the argument):
\begin{inparaenum}
 \item the point of the subject has to be equal to the reference point of the argument,
 \item the body of the subject has to be equal to the reference body of the argument,
  \item the body of the subject has to be equal to the body of the argument,
\end{inparaenum}
The figures below how the Xtext (Figure~\ref{fig:errorXtextEditor}) and Prolog-based (Figure~\ref{fig:errorPrologEditor}) editor react on a mistake on the first constraint in the above list. As shown in the figures they both provide information on the kind of error.

\section{Discussion}
\label{sec:discussion}

\subsection{Xcore versus Prolog DSL}
In this section we want to highlight some advantages and disadvantages of the Xcore and Prolog DSL for two use cases: the DSL developer and the DSL user.

For the \textbf{DSL user} both the external Xcore DSL and internal Prolog DSL currently provide an editor that offers checking of the constraints defined in the DSL.
However, because the Xcore DSL is easy to integrate into Eclipse, it immediately opens the way to all the tooling available in Eclipse. An example is the nice Xtext editor for the M1 level that can be generated in Eclipse from the Xcore DSL. 
Since, the Xcore DSL is however basically a text model, it is still possible to create any other parser or editor. Therefore, the Xcore DSL does not create a hard dependency on Eclipse or Xtext, while the tools of Eclipse and Xtext can still be used when desired.
The tooling around Prolog is not as developed as around Eclipse. Therefore, we implemented a simple editor for the M1 level ourselves. While the editor only offers basic checking and simple error reporting, it provides all the basic functionality to check the constraints defined in the DSL. 
The Prolog DSL has the extra advantage that uses the Prolog language, which is already executable. Therefore, it is more easy to create executable (Prolog) code from the M1 models defined using the Prolog DSL.


%

A \textbf{DSL developer} has to adapt or extend the external Xcore DSL and/or the internal Prolog DSL. The involved syntax of the OCL constraints make the Xcore DSL harder to `read'. Therefore, if the readability is an issue, it could be decided to natively implement the constraints in Xcore rather than using the OCL constraints. This is feasible in this case since we only use a small subset of the available OCL constraints. 
Since the Prolog syntax is quite intuitive it makes the internal Prolog DSL easier to `read'.


\subsection{Code generation: from M1 to M0}
An important limitation so far is that we have no code-generation support, i.e. the automatic transformation from the M1 to the M0 level is lacking.
In the robotics context this is an important limitation, since we need to obtain executable code. Moreover, preferably we want support for different programming languages (C++, python, \dots) and execution on different types of hardware (FPGA, normal PC, \ldots).
Therefore in future work we will also look at tools as Epsilon that allow to generate executable code.


\subsection{Future in robotics}
To ensure a future in robotics, not only code generation for the geometric semantics DSL is essential. Moreover, we need better support to write \textbf{entire robotic applications} at the M1 level. In robotics the code typically originates from different domains: geometry, kinematics, dynamics, state machines, estimators, etc. Therefore, it should be possible to write code that interleaves different DSLs.
To this end, different DSL (geometric semantics DSL, component models, kinematic and dynamic algorithms DSL~\cite{FrigerioBuchliCaldwell2011}, state charts~\cite{Klotzbuecher2010-DYROS}, motion specification DSL~\cite{Klotzbuecher2011-IROS}), \ldots) have to be supported \underline{at the same time}.
Tools will have to be developed that are \textbf{composable}, such that it possible to, depending on the application, load the relevant DSLs and to generate code from M1 specifications that are interleaving code of different DSLs.
Finally, the entire robotic application developed at M1 level has to be transformed to executable code (M0 level).


\section{Conclusion}
\label{sec:conclusion}
In this paper we presented both an external DSL in Xcore and an internal DSL in Prolog for geometric relations between rigid bodies such as relative position, orientation, pose, linear velocity, angular velocity, and twist founded on the geometric semantics~\cite{DeLaet-ram2012a,DeLaet-rrfc2012}.
These DSLs advance with respect to the available implementation in the general-purpose programming language (C++) by formalizing the underlying model of the geometric semantics.
Furthermore, we showed that these DSLs are the basis tools that assist the robot programmers and application developers. In an example we showed how editors built on top of the DSLs automatically check the semantic correctness of geometric operations on rigid-body coordinate representations while writing and editing the code. Furthermore, it was shown that these editors produce meaningful error statements when semantic constraints are violated.
We listed our experiences from writing the DSL up to using the editors. Finally, we discussed some things that are still lacking to integrate the geometric semantics DSL into the work flow of a robot programmer or application developer.

We believe that this paper has shown that the geometric semantics, due to is mature but concise nature, is an excellent example for the development of DSLs in robotics and the use of these DSLs in the work flow of a robot programmer or application developer.

 \section*{Acknowledgements}
 \small
All authors gratefully acknowledge the financial support by KU Leuven's
Concerted Research Action GOA/2010/011, 
European FP7 project Rosetta (2008-ICT-230902), 
European FP7 project BRICS (2008-ICT-231940), 
European FP7 project RoboHow (FP7-ICT-288533). 
Tinne De Laet is a Postdoctoral
Fellow of the Fund for Scientific Research--Flanders (F.W.O.) in Belgium.
\normalsize
\input{document.bbl}


\end{document}

%% file: macroGeometricPrimitives.tex
\newcommand{\refframe}[1]
{ \ifthenelse{\equal{#1}{}}{}{ \left\{\MakeLowercase{#1}\right\} \hspace{-0.2ex} } }

\newcommand{\orientation}[1]
{ \ifthenelse{\equal{#1}{}}{}{ \left[\MakeLowercase{#1}\right] \hspace{-0.1ex}} }

\newcommand{\point}[1]
{\MakeLowercase{#1}}

\newcommand{\body}[1]
{\mathcal{\MakeUppercase{#1}}\mathnormal{}}

\newcommand{\axis}[1]
{\MakeUppercase{#1}\mathnormal{}}

\newcommand{\fixedTo}[2]
{
#1
\ifthenelse{\equal{#2}{}}{}{
\hspace{-0.2ex}|\hspace{-0.2ex}
}
\body{#2}
}

%% file: macroSemantics.tex
\newcommand{\Orientation}[5]
{\textrm{Orientation\ifthenelse{\equal{#5}{}}{}{Coord}}\left(\fixedTo{\orientation{#1}}{#2}, \fixedTo{\orientation{#3}}{#4} \ifthenelse{\equal{#5}{}}{}{, \orientation{#5}} \right)}

\newcommand{\Position}[5]
{\textrm{Position\ifthenelse{\equal{#5}{}}{}{Coord}}\left(\fixedTo{\point{#1}}{#2} , \fixedTo{\point{#3}}{#4} \ifthenelse{\equal{#5}{}}{}{,\orientation{#5}} \right)}

\newcommand{\Pose}[7]
{\textrm{Pose\ifthenelse{\equal{#7}{}}{}{Coord}}\left(
\fixedTo{
\ifthenelse{\equal{#1}{#2}}{\refframe{#1}}{\left(
 \point{#1}, \orientation{#2} \right)} }{#3}
 , 
 \fixedTo{
 \ifthenelse{\equal{#5}{}}{ \ifthenelse{\equal{#4}{}}{}{\point{#4}}} { \ifthenelse{\equal{#4}{#5}}{\refframe{#4}}{\left(\point{#4}, \orientation{#5}\right)}}
 }
 {#6}
 \ifthenelse{\equal{#7}{}}{}{,\orientation{#7}} \right)}
 
\newcommand{\AngularVelocityBasic}[6]
{\textrm{AngularVelocity\ifthenelse{\equal{#5}{}}{}{Coord}}\left(
\ifthenelse{\equal{#1}{}}{\body{#2}}{
\ifthenelse{\equal{#6}{1}}{\forall}{}
\fixedTo{\orientation{#1}}{#2}
}
,
\ifthenelse{\equal{#3}{}}{\body{#4}}{
\ifthenelse{\equal{#6}{1}}{\forall}{}
\fixedTo{\orientation{#3}}{#4}
}  \ifthenelse{\equal{#5}{}}{}{,\orientation{#5}} \right)}

\newcommand{\AngularVelocity}[5]
{\AngularVelocityBasic{#1}{#2}{#3}{#4}{#5}{1}}

\newcommand{\LinearVelocityBasic}[6]
{\textrm{LinearVelocity\ifthenelse{\equal{#5}{}}{}{Coord}}\left(   \fixedTo{\point{#1}}{#2} , 
\ifthenelse{\equal{#3}{}}{\body{#4}}{\ifthenelse{\equal{#6}{1}}{\forall}{} 
\fixedTo{\point{#3}}{#4}} \ifthenelse{\equal{#5}{}}{}{,\orientation{#5}} \right)}

\newcommand{\LinearVelocity}[5]
{\LinearVelocityBasic{#1}{#2}{#3}{#4}{#5}{1}}

\newcommand{\VelocityBasic}[8]
{\textrm{Twist\ifthenelse{\equal{#7}{}}{}{Coord}}\left(
\fixedTo{
\ifthenelse{\equal{#2}{}}{\point{#1}}{
\ifthenelse{\equal{#1}{#2}}{\refframe{#1}}{\left(
 \point{#1}, \ifthenelse{\equal{#8}{1}}{\forall}{} \orientation{#2}\right)} }
 }{#3}
 , 
 \ifthenelse{\equal{#4}{}}{\body{#6}}{
 \fixedTo{
 \ifthenelse{\equal{#8}{1}}{\forall}{}
\ifthenelse{\equal{#5}{}}{\point{#4}}{
\ifthenelse{\equal{#4}{#5}}{\refframe{#4}}{\left(
 \point{#4},  \orientation{#5}\right)} }
 }{#6}
 }
 \ifthenelse{\equal{#7}{}}{}{,\orientation{#7}} \right)}

\newcommand{\Velocity}[7]
{\VelocityBasic{#1}{#2}{#3}{#4}{#5}{#6}{#7}{1}}

\newcommand{\ForceBasic}[3]
{\textrm{Force\ifthenelse{\equal{#3}{}}{}{Coord}}\left(
\body{#1},\body{#2}
\ifthenelse{\equal{#3}{}}{}{,\orientation{#3}} \right)}

\newcommand{\TorqueBasic}[4]
{\textrm{Torque\ifthenelse{\equal{#4}{}}{}{Coord}}\left(   \fixedTo{\point{#1}}{#2} , 
\body{#3}
\ifthenelse{\equal{#4}{}}{}{,\orientation{#4}} \right)
}

\newcommand{\WrenchBasic}[4]
{\textrm{Wrench\ifthenelse{\equal{#4}{}}{}{Coord}}\left(
\fixedTo{\point{#1}}{\body{#2}}, \body{#3}
 \ifthenelse{\equal{#4}{}}{}{,\orientation{#4}} \right)}


%% file: macroSemanticOperations.tex
\newcommand{\changePoint}[2]
{#1.\textrm{changePoint}\left(#2\right)
}

\newcommand{\changePointPosition}[2]
{\changePoint{#1}{#2}
}

%% file: listingsSettings.tex
\lstdefinelanguage{pseudoCode}
{
keywordstyle=\color[rgb]{0.627,0.126,0.941},
morekeywords={Pose,PoseSemantics,PoseCoordinates,PoseCoordinatesSemantics,
Orientation,OrientationSemantics,OrientationCoordinates,OrientationCoordinatesSemantics,
Position,PositionSemantics,PositionCoordinates,PositionCoordinatesSemantics,
Twist,TwistSemantics,TwistCoordinates,TwistCoordinatesSemantics,
AngularVelocity,AngularVelocitySemantics,AngularVelocityCoordinates,AngularVelocityCoordinatesSemantics,
LinearVelocity,LinearVelocitySemantics,LinearVelocityCoordinates,LinearVelocityCoordinatesSemantics,
Force,ForceSemantics,ForceCoordinates,ForceCoordinatesSemantics,
Torque,TorqueSemantics,TorqueCoordinates,TorqueCoordinatesSemantics,
Wrench,WrenchSemantics,WrenchCoordinates,WrenchCoordinatesSemantics},
sensitive=false,
morecomment=[l]{//},
emph={Result},
emphstyle=\color[rgb]{0.627,0.126,0.941},
emph={[2]Semantic,output},
emphstyle={[2]\color[rgb]{0.74,0.08,0.32}},
emph={[3]NOK},
emphstyle={[3]\color{red}},
emph={[4]OK},
emphstyle={[4]\color[rgb]{0.2,0.6,0.2}},
morestring=[b]``,
keywordstyle=\color[rgb]{0,0.63,0.69},
}

\lstdefinelanguage{xcore}
{
keywordstyle=\color[rgb]{0.627,0.126,0.941},
morekeywords={class,extends,String,abstract,refers,@Pivot,derivation,self,derived},
sensitive=false,
morecomment=[l]{*},
emph={Result},
emphstyle=\color[rgb]{0.627,0.126,0.941},
emph={[2]Semantic,output},
emphstyle={[2]\color[rgb]{0.74,0.08,0.32}},
emph={[3]NOK},
emphstyle={[3]\color{red}},
emph={[4]OK},
emphstyle={[4]\color[rgb]{0.2,0.6,0.2}},
morestring=[b]``,
keywordstyle=\color[rgb]{0,0.63,0.69},
}

\lstnewenvironment{xcoreEnv}[1][]
  {\lstset{
  	language=xcore,
  	label=#1,
	inputencoding=utf8x,
	backgroundcolor=\color{lbcolor},
	tabsize=2,
	rulecolor=,
        basicstyle=\footnotesize \ttfamily,
        upquote=true,
        aboveskip={1.0\baselineskip},
        columns=fixed,
        showstringspaces=false,
        extendedchars=true,
        breaklines=true,
        prebreak = \raisebox{0ex}[0ex][0ex]{\ensuremath{\hookleftarrow}},
        showtabs=false,
        showspaces=false,
        showstringspaces=false,
        identifierstyle=\ttfamily,
        keywordstyle=\color[rgb]{0,0,1},
        commentstyle=\color[rgb]{0.133,0.545,0.133},
        stringstyle=\color[rgb]{0,0,0},
         keywordstyle=\color[rgb]{0.627,0.126,0.941}
,
}
}
{}

\lstnewenvironment{prologEnv}[1][]
  {\lstset{
  	language=Prolog,
  	label=#1,
	inputencoding=utf8x,
	backgroundcolor=\color{lbcolor},
	tabsize=2,
	rulecolor=,
        basicstyle=\footnotesize \ttfamily,
        upquote=true,
        aboveskip={1.0\baselineskip},
        columns=fixed,
        showstringspaces=false,
        extendedchars=true,
        breaklines=true,
        prebreak = \raisebox{0ex}[0ex][0ex]{\ensuremath{\hookleftarrow}},
        showtabs=false,
        showspaces=false,
        showstringspaces=false,
        identifierstyle=\ttfamily,
        keywordstyle=\color[rgb]{0,0,1},
        commentstyle=\color[rgb]{0.133,0.545,0.133},
        stringstyle=\color[rgb]{0,0,0},
         keywordstyle=\color[rgb]{0.627,0.126,0.941}
,
}
}
{}

\lstnewenvironment{cPlusPlusEnv}[1][]
  {\lstset{
  	language=c++,
  	label=#1,
	inputencoding=utf8x,
	backgroundcolor=\color{lbcolor},
	tabsize=2,
	rulecolor=,
        basicstyle=\footnotesize \ttfamily,
        upquote=true,
        aboveskip={1.0\baselineskip},
        columns=fixed,
        showstringspaces=false,
        extendedchars=true,
        breaklines=true,
        prebreak = \raisebox{0ex}[0ex][0ex]{\ensuremath{\hookleftarrow}},
        showtabs=false,
        showspaces=false,
        showstringspaces=false,
        identifierstyle=\ttfamily,
        keywordstyle=\color[rgb]{0,0,1},
        commentstyle=\color[rgb]{0.133,0.545,0.133},
        stringstyle=\color[rgb]{0,0,0},
        keywordstyle=\color[rgb]{0.627,0.126,0.941},
morekeywords={Pose,PoseSemantics,PoseCoordinates,PoseCoordinatesSemantics,
Orientation,OrientationSemantics,OrientationCoordinates,OrientationCoordinatesSemantics,
Position,PositionSemantics,PositionCoordinates,PositionCoordinatesSemantics,
Twist,TwistSemantics,TwistCoordinates,TwistCoordinatesSemantics,
AngularVelocity,AngularVelocitySemantics,AngularVelocityCoordinates,AngularVelocityCoordinatesSemantics,
LinearVelocity,LinearVelocitySemantics,LinearVelocityCoordinates,LinearVelocityCoordinatesSemantics,
Force,ForceSemantics,ForceCoordinates,ForceCoordinatesSemantics,
Torque,TorqueSemantics,TorqueCoordinates,TorqueCoordinatesSemantics,
Wrench,WrenchSemantics,WrenchCoordinates,WrenchCoordinatesSemantics,
PoseGraph,TwistGraph,
GetPose,GetTwist,
AddPose,AddTwist
},
}
}
{}

\lstnewenvironment{cPlusPlusEnvOneColumn}[1][]
  {\lstset{
  	language=c++,
  	label=#1,
	inputencoding=utf8x,
	backgroundcolor=\color{lbcolor},
	tabsize=2,
	rulecolor=,
        basicstyle=\footnotesize \ttfamily,
        upquote=true,
        aboveskip={1.0\baselineskip},
        columns=fixed,
        showstringspaces=false,
        extendedchars=true,
        breaklines=true,
        prebreak = \raisebox{0ex}[0ex][0ex]{\ensuremath{\hookleftarrow}},
        showtabs=false,
        showspaces=false,
        showstringspaces=false,
        identifierstyle=\ttfamily,
        keywordstyle=\color[rgb]{0,0,1},
        commentstyle=\color[rgb]{0.133,0.545,0.133},
        stringstyle=\color[rgb]{0,0,0},
        keywordstyle=\color[rgb]{0.627,0.126,0.941},
morekeywords={Pose,PoseSemantics,PoseCoordinates,PoseCoordinatesSemantics,
Orientation,OrientationSemantics,OrientationCoordinates,OrientationCoordinatesSemantics,
Position,PositionSemantics,PositionCoordinates,PositionCoordinatesSemantics,
Twist,TwistSemantics,TwistCoordinates,TwistCoordinatesSemantics,
AngularVelocity,AngularVelocitySemantics,AngularVelocityCoordinates,AngularVelocityCoordinatesSemantics,
LinearVelocity,LinearVelocitySemantics,LinearVelocityCoordinates,LinearVelocityCoordinatesSemantics,
Force,ForceSemantics,ForceCoordinates,ForceCoordinatesSemantics,
Torque,TorqueSemantics,TorqueCoordinates,TorqueCoordinatesSemantics,
Wrench,WrenchSemantics,WrenchCoordinates,WrenchCoordinatesSemantics,
PoseGraph,TwistGraph,
GetPose,GetTwist,
AddPose,AddTwist
},
}
}
{}

\lstnewenvironment{consoleOutputEnv}[1][]
  {\lstset{
  	language=consoleOutput,
  	label=#1,
	inputencoding=utf8x,
	backgroundcolor=\color{lbcolor},
	tabsize=2,
	rulecolor=,
        basicstyle=\footnotesize \ttfamily,
        upquote=true,
        aboveskip={1.0\baselineskip},
        columns=fixed,
        showstringspaces=false,
        extendedchars=true,
        breaklines=true,
        prebreak = \raisebox{0ex}[0ex][0ex]{\ensuremath{\hookleftarrow}},
        showtabs=false,
        showspaces=false,
        showstringspaces=false,
        identifierstyle=\ttfamily,
        commentstyle=\color[rgb]{0.133,0.545,0.133},
        stringstyle=\color[rgb]{0.627,0.126,0.941},
}
}
{}

\lstnewenvironment{consoleOutputEnvOneColumn}[1][]
  {\lstset{
  	language=consoleOutput,
  	label=#1,
	inputencoding=utf8x,
	backgroundcolor=\color{lbcolor},
	tabsize=2,
	rulecolor=,
        basicstyle=\footnotesize \ttfamily,
        upquote=true,
        aboveskip={1.0\baselineskip},
        columns=fixed,
        showstringspaces=false,
        extendedchars=true,
        breaklines=true,
        prebreak = \raisebox{0ex}[0ex][0ex]{\ensuremath{\hookleftarrow}},
        showtabs=false,
        showspaces=false,
        showstringspaces=false,
        identifierstyle=\ttfamily,
        commentstyle=\color[rgb]{0.133,0.545,0.133},
        stringstyle=\color[rgb]{0.627,0.126,0.941},
}
}
{}
\usepackage{caption}
\usepackage{courier}
\usepackage{etoolbox}

\DeclareCaptionFont{white}{\color{white}}
\DeclareCaptionFormat{listing}{\colorbox[cmyk]{0.43, 0.35, 0.35,0.01}{\parbox{\columnwidth}{\hspace{0pt}#1#2#3}}}